\documentclass[manuscript,screen]{acmart}
\usepackage{subfigure}
\AtBeginDocument{%
  \providecommand\BibTeX{{%
    \normalfont B\kern-0.5em{\scshape i\kern-0.25em b}\kern-0.8em\TeX}}}

\begin{document}

\setcopyright{acmcopyright}
\acmJournal{TALLIP}
\acmYear{2020} \acmVolume{1} \acmNumber{1} \acmArticle{1} \acmMonth{1} 
\acmPrice{15.00}
\acmDOI{10.1145/3434239}

\title{geoGAT: Graph Model Based on Attention Mechanism for Geographic Text Classification}

\author{Weipeng Jing}
\email{jwp@nefu.edu.cn}
\authornotemark[1]
\affiliation{
  \institution{Northeast Forestry University}
  \city{Harbin}
  \country{China}
  \postcode{150040}
}

\author{Xianyang Song}
\email{sxy56713@nefu.edu.cn}
\affiliation{
  \institution{Northeast Forestry University}
  \city{Harbin}
  \country{China}
  \postcode{150040}
}

\author{Donglin Di}
\email{donglin.ddl@gmail.com}
\affiliation{%
  \institution{Tsinghua University}
  \city{Haidian Qu}
  \state{Beijing Shi}
  \country{China}}
  
\author{Houbing Song}
\email{SONGH4@erau.edu}
\affiliation{%
  \institution{Department of Electrical Engineering and Computer Science, Embry-Riddle Aeronautical University}
  \city{Daytona Beach}
  \state{FL}
  \country{USA}
  \postcode{32114}}


\begin{abstract}
  In the area of geographic information processing. There are few researches on geographic text classification. However, the application of this task in Chinese is relatively rare. In our work, we intend to implement a method to extract text containing geographical entities from a large number of network text. The geographic information in these texts is of great practical significance to transportation, urban and rural planning, disaster relief and other fields. We use the method of graph convolutional neural network with attention mechanism to achieve this function. Graph attention networks is an improvement of graph convolutional neural networks. Compared with GCN, the advantage of GAT is that the attention mechanism is proposed to weight the sum of the characteristics of adjacent nodes. In addition, We construct a Chinese dataset containing geographical classification from multiple datasets of Chinese text classification. The Macro-F Score of the geoGAT we used reached 95\% on the new Chinese dataset.
\end{abstract}

\ccsdesc[500]{Information systems~Clustering and classification}
\ccsdesc[300]{Computing methodologies~Information extraction}
\keywords{graph neural networks, attention mechanism, toponym recognition, text classification}

\maketitle

\section{Introduction}
Toponym recognition and toponym resolution, extraction of geographic relations and geolocation are all simple and important tasks in geographic text knowledge mining. Many researchers applied deep learning on accomplish those tasks. In most of the current researches on geographic text knowledge mining, those tasks are usually combined with similar tasks in natural language processing (NLP). It is a time-consuming but valuable work to find the text with geographic information among numerous texts. Some kind of violence or natural disaster often happens in our life, people would like to usually send text messages on social media to ask for help. Because social media networks are time-sensitive, lots of information can be easily passed through. It is difficult to scan messages directly to determine where people need assistance. Therefore, if there is a way to quickly separate geographically related texts from other texts, relevant departments are able to better carry out rescue actions by analyzing relevant contents.

Text classification is a basic assignment in natural language processing. Its application in daily life is ubiquitous. The most common applications are intention analysis, emotion recognition, sentence matching and question and answer matching. On account of the rapid development of neural network technology in recent decades, its application in various fields is mature, more and more researchers pay attention to the neural networks.  
Deep learning based on neural network is widely applied, among which the methods include convolutional neural networks (CNN)\cite{726791}\cite{kim2014convolutional}, recurrent neural networks (RNN)\cite{lipton2015critical} and long short-term memory (LSTM)\cite{doi:10.1162/neco.1997.9.8.1735}. LSTM belongs to the variant form of RNN. Due to the memory property of LSTM, this kind of neural networks has advantages in processing sequence correlation data such as text. 
In today's Internet, there are substantial text data of network structure, such as social network, knowledge map. They have non-Euclidean structures.
However, the traditional neural network is not good at processing non-Euclidean data, and it is difficult to extract valuable information from it. Under the impetus of multiple factors, researchers used the ideas of convolutional network, circulating network and depth automatic encoder for reference, defined and designed the neural network structure for processing graph data. A new model of deep learning has come to be known as the graph convolutional neural network. GCN also has advantages compared with the traditional neural network. The above CNN and RNN cannot process the feature representation of graph embedding in non-sequential order. GCN is propagated on each node separately, ignoring the order of input between nodes. In other words, the output of GCN is not shifted with the input order of the nodes.

In this work, we propose geoGAT: geographical graph attention network which is able to identify text containing geographical knowledge in the text. This method is based on the graph attention networks (GAT)\cite{velikovi2017graph}. Graph attention networks is formed by the attention-introducing mechanism of graph convolutional network (GCN)\cite{kipf2016semisupervised}. Attention mechanism is applied on graph convolutional network to better integrate the correlation of vertex features into the model\cite{8294302}. GAT compensates for the shortcomings of GCN. GCN gives the same weight to adjacent nodes in the same-order neighborhood of nodes, which limits the ability of the model to capture spatial information correlation.
We first process the Chinese geo-text into a graph that can be processed by the graph neural network. Then the graph data is input into the geoGAT by converting the graph data into matrix format and classified by the output feature vectors. This allows the text with geographic information to be identified from whole Chinese dataset. To sum up, the main contributions of our work are as follows:
\begin{itemize}
\item  We propose a novel method for geographic text
classification. geoGAT enhances the ability to capture the correlation of spatial information by using an attention mechanism instead of a standardized operation in GCN to collect and derive the characteristic representation of adjacent nodes and aims to extract geographic information from network text as accurately as possible, so that various geographic information can be applied in other fields.
\item By collecting texts in different Chinese datasets, a Chinese dataset containing geographical texts is constructed and the corpus in the dataset is divided into five categories. Each piece of data is segmented by character level and word level respectively. By training dataset in different models, the ability of each model to extract geographic information from the data is analyzed.
\end{itemize}

\section{Related Work}
\subsection{Study of Geographic Text}

Text has widely occupied a large proportion in network information due to its small capacity, easy to edit and strong expressive ability. Therefore, the emergence and development of natural language processing is to obtain the knowledge and information people need from the text and it is particularly necessary to summarize and classify the text.

The geographical text occupies a certain proportion among the network text and there are few study on geographic text information. In geographic text research, the information related to geography is recognized by algorithms or models to separate it from other texts. This separated text with geographic information is summarized as geo-txt. The geo-text obtained by separation has a wide range of applications\cite{hu_geo-text_2018}. The main research focuses on a kind of geo-txt: toponyms recognition\cite{lieberman_multifaceted_2011}. The primary goal is to find the nouns or phrases that can represent geographical entities in the text. Another study is toponym resolution. There is an interesting Chinese place name \textit{JiangNan}, can refer to a famous novelist who has the same pen name. That main goal is to use disambiguation and geolocation methods to determine the specific meaning contained in the text\cite{buscaldi_conceptual_2008}\cite{doi:10.1080/13658810701626236}. Geographic disambiguation is a necessary to other application about Geographic Information System (GIS). 

There's another kind of interesting text that can be sort of geo-txt, most of text is written by people. Therefore, the text often contains people's views and emotions towards something. This kind of text has not only the nature of geography, but also people's subjective assumption. It is of great significance for the evaluation and judgment of a certain area. The emotion analysis is part of NLP, and many researchers have applied it to geographic information. We can get the distribution of people's evaluation of some scenic spots through the analysis of a large amount of data, so as to get the satisfaction degree of this scenic spot among tourists.

In addition to the above applications, geo-txt is also available from social media. There is a common situation. In the event of various accidents in our lives, people release information for help through social media. By locating the geographical elements in these social information, rescue operations can be carried out more timely\cite{doi:10.1111/tgis.12627}. In 2018, researchers conducted geolocation of users through text and network context\cite{rahimi2018semisupervised}, using the currently popular graph convolutional network.

\subsection{Methods of Text Classification}
The task of text classification is to automatically classify text according to a certain classification system and standard and to find the relationship between the characteristics of documents and the types of documents on the basis of labeled documents.

At the end of 1950s, The idea of using word frequency statistics for automatic text classification was first proposed. In 1960, M.E.Maron proposed a new search member (keyword) for classification task, making automatic classification technology enter a new era\cite{10.1145/321033.321035}. At the end of the 20th century, text classification was mainly processed based on knowledge engineering. This method is to form a classification system based on rules manually written by experts in related fields. This approach has significant limitations.
Traditional knowledge engineering classification has been unable to meet the needs of information industry. Most of that are knowledge-based and statistics-based. It mainly uses feature engineering and classification algorithm, and feature engineering determines the upper limit of classification effect\cite{rousseau_text_2015}. Text feature engineering is divided into three parts: text preprocessing, feature extraction and text representation.
The ultimate goal is to convert text into a format that a computer can understand and encapsulate enough information to categorize. Naive Bayes classification algorithm, KNN algorithm and support vector machine algorithm (SVM)\cite{SANCHEZA20035} are commonly used in classifiers.

The traditional text classification of Chinese text has high representation dimension and high sparse degree. The reason why the text representation method cannot be applied in deep learning is that method cannot express the features well. If we want to solve the problem of text classification, it suppose to solve the problem of representation of text characteristics first.

There are many kinds of distributed representation of text. The earliest method is one-hot encoding. There's only one dimension in vector space that is 1, and the rest of it is 0. Then word embedding is well known to public and be extensively used when the Google invented Word2Vec\cite{mikolov2013efficient}\footnote{https://code.google.com/archive/p/word2vec/}. It is able to learn the representation of text through the context of text. It indicated that the way in which words are embedded determines the effect of neural networks in natural language processing.

Some recent studies have achieved many great classification effects\cite{NIPS2013_5021}\cite{pennington_glove_2014} by means of a large degree of optimization in word embedding. Another way to achieve good results in text classification is to optimize the training model. In 2016, the author of Word2Vec published a paper proposing the FastText\cite{arm2016bag} which applied average word-embedding followed by softmax layer to achieve some simple tasks.
Yoon Kim proposed TextCNN in 2014\cite{kim2014convolutional}. As a classic neural network model, CNN has the characteristics of fast running speed and strong feature extraction ability. Therefore, researchers applied CNN to the task of natural language processing after the great success of CNN in the field of computer vision. DCNN\cite{Huang_2017_CVPR} uses a dynamic pooling method to conduct semantic modeling for sentences. Later, RCNN\cite{Girshick_2014_CVPR} was proposed. The author removed the convolution kernel structure in CNN and used RNN to capture the context features of words.
\subsection{GNN}
Graph neural networks (GNN) has recently caught the attention of researchers. The representation of the graph neural network was proposed by Gori et al in 2005\cite{1555942}. Further illustrated by Scarselli et al\cite{4773279} in 2009.
In the early study of graph neural networks, the information of neighbor nodes was propagated through a circular method until a stable fixed point was reached to learn the representation of target nodes. Inspired by CNN's success in the field of computer vision, graph convolutional networks was proposed.

In 2013, Bruna designed a variant of graph convolution for the first time based on the spectral theory. Later, Battaglia et al took the location map network as the building block to learn relational data and used a unified framework to review some neural networks\cite{battaglia2018relational}. Unlike other networks, graph neural networks retain a state that represent information from any depth in their neighborhood.
The development of network architecture and parallel computing solved problem that the original GNN was difficult to train for fixed points, 

In recent years, models based on graph convolutional networks (GCN) and gated graph neural networks (GGNN)\cite{li2015gated} have demonstrated breakthrough performance in many research fields. Lee et al conducted a partial investigation of the graph attention model\cite{lee2018attention}. Bengio's team proposed graph attention networks (GAT), a network architecture based on neighbor-node attention mechanism, which is used to handle computational graph of complex, irregular structures and achieved industry-best results in three difficult benchmarks\cite{velikovi2017graph}. The researchers show the model has the potential to handle arbitrary irregular structure graph in the future. After that, the graph neural network will slow down when the graph data is complex. In 2018, IBM proposed FastGCN to achieve fast learning by means of importance sampling and graph convolutional network\cite{chen2018fastgcn}. In terms of text classification, textGCN is proposed\cite{yao_graph_2018}.
This model solves the co-occurrence of global words ignored by CNN, RNN and other models. A novel text classification method based on graph neural network is used to generate graph embedded model.

\section{Methodology}
\begin{figure*}
    \centering
    \includegraphics[width=0.8\textwidth]{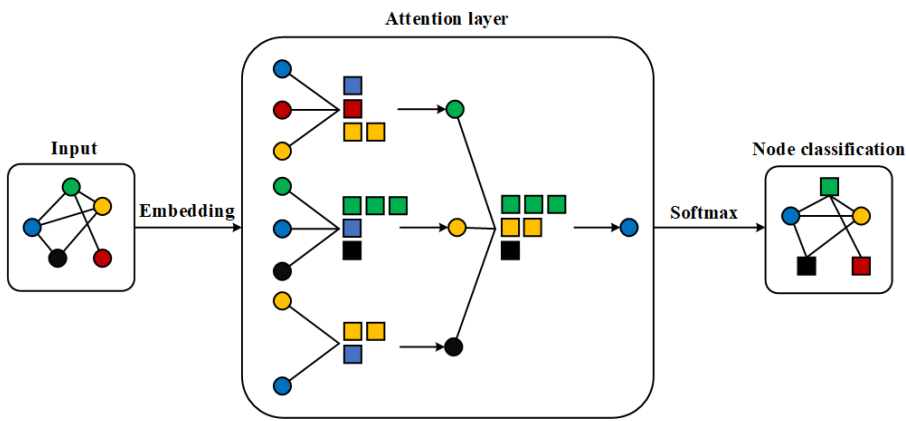}
    \caption{Overview of the model. The text at the input layer is converted into a heterogeneous graph. The conversion method is shown in Figure \ref{fig:text}. In the graph, different colored nodes represents words and sequences. Through the attention layer, the number of squares of different colors represents the magnitude of different attention coefficients. The attention weight of each node is calculated by its adjacent node, so as to predict the output characteristics of each node. The nodes are classified through softmax layer in the end. Squares and circles describe how nodes fall into different classes.}
    \label{fig:overview2}  
\end{figure*}

\begin{figure}
    \centering
    \includegraphics[width=0.65\textwidth]{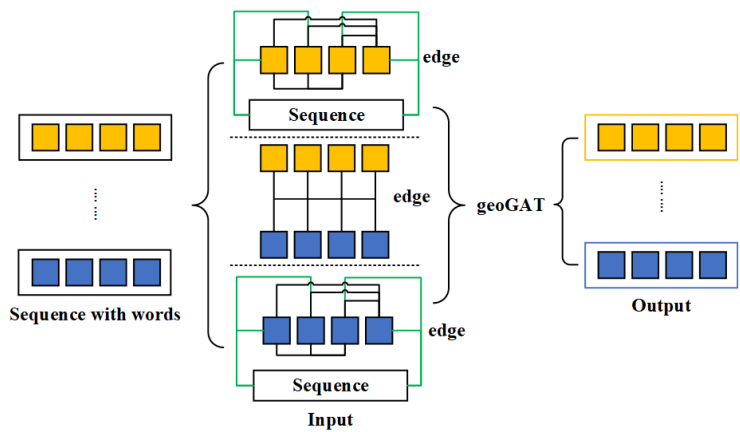}
    \caption{Text is converted to a graph. Each of the large rectangles on the left represents a sequence of text and the small squares in the large rectangle represent the words in the sequence.
The middle section describes the constructed graph as the input to the model, and the nodes of the graph are composed of words and sequences.
There are two kinds of edges, the black edge represents the edge between word and word, and the green edge represents the edge between word and the corresponding sequence. The right of the figure indicates that each sequence belongs to a different category after the input is passed through the model.}
    \label{fig:text}
\end{figure}

In the field of image recognition, the object to be operated is an image with two-dimensional structure, so the convolution kernel in CNN is used to extract features by moving on the image. In natural language processing, the manipulated object is a sequence with a one-dimensional structure, and the structure of RNN learn the relationship between the information before and after the sequence superior, so as to better capture the characteristics of the sequence. Data such as images and text have a regular structure, but there are also many data with irregular structure, such as social network data, graph, knowledge graph and chemical formula. Therefore, GCN is designed to extract features from the graph structure data. 

The overview of the model is described in the Figure \ref{fig:overview2}. The model on the left of the figure shows the input layer of the model. In order to use the data as input to the graph model, the one-dimensional geographic text needs to be transformed into the graph structure.
The intermediate module shows the graph data passing through the attention layer. The attention mechanism predicts the output characteristics of each node by analyzing the input characteristics of each node. The right module is the model after training, and the node features are obtained and classified through softmax layer.

Figure \ref{fig:text} describes the detailed process of converting text into a graph structure.
An undirected graph is made up of nodes and edges. $Sequence$ represents each piece of text in the dataset, each of these little squares represents each character or word in the text (depending on whether word-level segmentation or character-level segmentation are used) and each sequence and word constitutes the node of the entire graph. The line between squares in the figure pointing from word to word represents one kind of edge, while the line pointing from word to sequence represents another kind of edge. Using this method, the entire dataset is constructed into a large graph structure. There is a method for calculating the weight of every kind of edge. It is shown in eq \ref{eq7}. $A_{ij}$ is the weight of the edge, and $i,j$ is the word or sequence.
\begin{equation}
   A_{ij}=\left\{
   \begin{array}{lcl}
    PMI(i,j)        &   & i,j \text{ are words}                  \\
    TF-IDF_{ij}     &   & i \text{ is sequence},j \text{ is word}\\
    1               &   & i=j                                    \\
    0               &   & \text{otherwise}
   \end{array}
\right.
\label{eq7}
\end{equation}

TF-IDF is adopted to calculate the weight of edge between word and sequence. $TF$ represents Term Frequency, indicating the frequency of a word appearing in the text. $IDF$ represents the Inverse Document Frequency. $IDF$ of a certain word, which can be calculated by dividing the total number of texts by the number of texts containing this word by adding one (to prevent the denominator from being 0) and then logarithm of the obtained quotient. TF-IDF can be obtained by multiplying $TF$ by $IDF$.
The meaning of TF-IDF is that if a word or word increases in a direct proportion with its occurrence frequency in the text, but decreases in an inverse proportion with its occurrence frequency in the whole corpus.
This method can be used to evaluate the importance of a character or word to a particular text in the entire dataset. 

The weights of words edges are based on the global word co-occurrence information of the word. Word co-occurrence information used a fixed-size sliding window to slide statistical word co-occurrence information in the corpus and then point mutual information (PMI) is used to calculate the weight of the line between the two word nodes. The eq \ref{eq8},\ref{eq9},\ref{eq10} describes the specific calculation process.
\begin{equation}
   PMI(i,j)=\log\frac{p(i,j)}{p(i)p(j)}
   \label{eq8}
\end{equation}

\begin{equation}
   p(i,j)=\frac{W(i,j)}{W}
   \label{eq9}
\end{equation}

\begin{equation}
   p(i)=\frac{W(i)}{W}
   \label{eq10}
\end{equation}

$W$ represents the total number of sliding windows, $W(i)$ represents the number of sliding windows containing word $i$ in the entire corpus and $W(i,j)$ represents the number of sliding windows containing both word $i$ and word $j$.
The value of PMI is calculated by eq \ref{eq8}. If the result is positive, it means that there is a strong semantic correlation between the two words; otherwise, it means that the semantic correlation is small or there is no semantic correlation and the weights are given to the two words with a positive PMI. Through the above operation, the corpus can be constructed into a graph structure.

\subsection{Graph Convolutional Network}
Suppose a batch of graph $G=(V,E)$ contains $N(|V|=n)$ nodes and each node has $M$ features, then an $N \times M$ feature matrix $X$ is formed between the nodes and their features. The relationship between $A$ node and other nodes constitutes the adjacency matrix $A$.
The feature matrix $X$ and the adjacency matrix $A$ constitute the input to GCN.
As a neural network, the core equation of GCN is as follows:

\begin{equation}
  H^{l+1}=\sigma(\Tilde{D}^{-\frac{1}{2}}\Tilde{A}\Tilde{D}^{-\frac{1}{2}}H^{(l)}W^{(l)})
  \label{eq_h}
\end{equation}

$\sigma$ is the activation function, $D$ is the diagonal node degree matrix, $\Tilde{A}=A+I$, where $I$ is the identity matrix and $\Tilde{D}$ is the diagonal node matrix of $\Tilde{A}$. $H$ is the feature matrix. In the state of $H_0$, $H$ is equal to $X$. $W$ is the parametric matrix.
Suppose we construct a two-layer GCN and the activation function uses ReLU and Softmax respectively, then the overall forward propagation is as follows:
\begin{equation}
  f(X,A)=softmax(AReLU(AXW^{(0)})W^{(1)})
  \label{eq1}
\end{equation}

$A$ represents $\Tilde{D}^{-\frac{1}{2}}\Tilde{A}\Tilde{D}^{-\frac{1}{2}}$ in eq \ref{eq1}. $X$ is the characteristic matrix of the graph, $W^{(0)}$ is the parameter matrix that needs training and $AXW^{(0)}$ is the representation of words and sequence nodes. $AReLU(AXW^{(0)})W^{(1)}$ is the second level representation of the node. The trained feature representation is fed to the softmax layer for classification.

\subsection{Graph Attention Mechanism}
Graph attention layer is introduced to graph attention network. Replacing the normalized function in GCN with a neighbor node feature aggregation function that uses attention weights. As shown in the eq \ref{eq2},\ref{eq3}. $h$ represents the characteristics of nodes, similar to $AXW^{(0)}$ in eq \ref{eq1}. In the attention layer, the input in the network is $N$ nodes and each node has $F$ characteristics. The output is $N$ nodes and each node has $F'$ features. 
\begin{figure}
    \centering
    \includegraphics[width=0.7\textwidth]{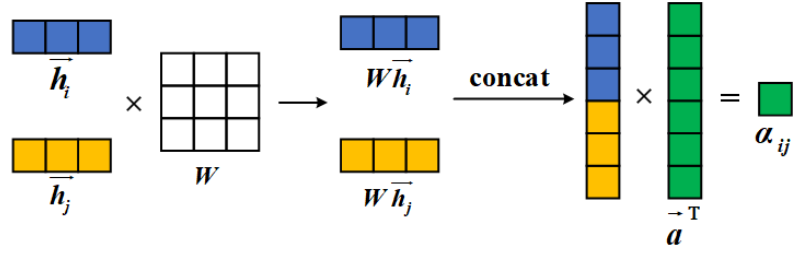}
    \caption{The process of calculating the coefficient of attention.
$\Vec{h}_i$ and $\Vec{h}_j$ in the figure represent the features of node $i$ and $j$. Assume that the node has a feature dimension of 3 ($F=3$). The dimension of the weight matrix $W$ is 3 by 3 ($F\times F'$). The output dimension multiplied by the input node features and weight matrix is $1\times F'$. Concat operation on $W\Vec{h}_i$ and $W\Vec{h}_j$ is multiplied by the attention kernel to obtain the attention coefficient $e_{ij}$ between the two nodes.}
    \label{fig:GAT-fomula3}
\end{figure}
\begin{equation}
  h = \{\Vec{h}_1,\Vec{h}_2,...,\Vec{h}_N\}, \Vec{h}_i\in \mathbb{R}^F
  \label{eq2}
\end{equation}
\begin{equation}
  h^{'} = \{\Vec{h^{'}_1},\Vec{h^{'}_2},...,\Vec{h^{'}_N}\}, \Vec{h^{'}_i}\in \mathbb{R}^{F'}
  \label{eq3}
\end{equation}
Our aim is to train all nodes to obtain a weight matrix $W\in R^{F\times F'}$. $W$ is the relationship between the $F$ features of the input and the $F'$ features of the output.
First introducing attention mechanism for each node where the attention coefficients is shown in eq \ref{eq4}. $i,j$ represents two different nodes. The meaning of $e_{ij}$ is how important node $j$ is to node $i$. In this way, each node gets its own weight of attention.
\begin{equation}
  e_{ij}=\Vec{a}^\mathrm{T}(W\Vec{h}_i,W\Vec{h}_j)
  \label{eq4}
\end{equation}
Then the method use the masked attention mechanism to allocate attention to $N_i$ among all neighbor nodes of node $i$. eq \ref{eq5} shows that $softmax$ is applied to regularize the adjacent node $j$ of $i$ to better calculate the attention coefficients.
\begin{equation}
  \alpha_{ij}=softmax(e_{ij})=\frac{exp(e_{ij})}{\sum_{k\in N_i} exp(e_{ik})}
  \label{eq5}
\end{equation}
Combine the eq \ref{eq4} and eq \ref{eq5}. Complete attention coefficient is defined in eq \ref{eq6}.
\begin{equation}
  \alpha_{ij}=\frac{{}\exp{(LeakyReLu(\Vec{a}^\mathrm{T}[W\Vec{h}_i||W\Vec{h}_j]))}}{\sum_{k\in N_i}\exp{(LeakyReLu(\Vec{a}^\mathrm{T}[W\Vec{h}_i||W\Vec{h}_k]))}}
  \label{eq6}
\end{equation}

\subsection{Multi-head Attention}
The attention mechanism is a single-layer feedforward neural network and $\Vec{a}$ is the weight matrix between the connecting layers. In order to better train the effect, $LeakyReLu$ is introduced into the output layer of the network. $e_{ij}$ in eq \ref{eq4} and $\alpha_{ij}$ in eq \ref{eq6} are attention coefficients and $\alpha_{ij}$ is normalized on the basis of $e_{ij}$. The above process can be described by Figure \ref{fig:GAT-fomula3}. 
Adding multi-head attention into the calculation formula is conducive to the learning of self-attention. The specific equation is as follows:
\begin{equation}
  \Vec{h'}_i=softmax(\frac{1}{K}\sum_{k=1}^K\sum_{j\in N_i}\alpha^k_{ij}W^k\Vec{h_j})
  \label{hi}
\end{equation}

$K$ is the number of attention mechanisms, $k$ is the k-th of $K$. The k-th attention coefficient of a node is expressed as $\alpha^k_{ij}$, the weight matrix under the k-th attention mechanism is expressed as $W^k$ and $softmax$ is the nonlinear activation function. So we get the transformation from $h$ to $h'$.

In the right-most module in Figure \ref{fig:overview2}, each node is classified by the softmax function that is described in eq \ref{softmax}. $h_i$ is the characteristic of a node as described earlier, $softmax_i$ represents the probability of belonging to this class and $C$ represents the number of classification categories. The probability of the node under each label can be obtained through the equation.
\begin{equation}
  softmax_i=\frac{e^{h_i}}{\sum_{k=1}^Ce^{a_k}}
  \label{softmax}
\end{equation}

\begin{figure}
    \centering
    \includegraphics[width=0.5\textwidth]{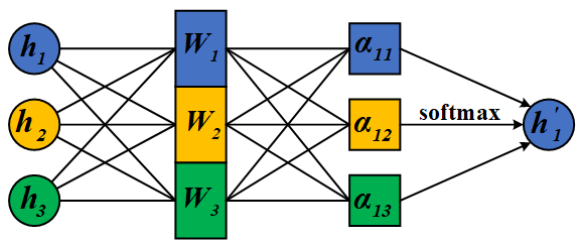}
    \caption{The computational process of the multi-head attention mechanism.
Assume that the two adjacent nodes of node $1$ are node $2$ and node $3$, and the number of multi-head attention mechanisms $K$ is equal to 3.
To compute the output features of $h_1$.
Each node is multiplied by the corresponding weight matrix $W_i$ and the attention coefficient $\alpha_{ij}$ and repeat 3 times $(K=3)$.
After the results of the three cycles are added and averaged, the output features are obtained through a nonlinear activation function $(softmax)$.}
    \label{fig:GAT-fomula4}
\end{figure}
\begin{table*}
\caption{The definition of the confusion matrix for Geographical classification}
\label{confusion matrix}
\begin{tabular}{c|cc}
\toprule
 & \textbf{Classify as Geography} & \textbf{Classify as Non-Geography }   \\
\midrule
\textbf{Geography}  & True Positive (\textit{TP})  & False Negative (\textit{FN}) \\
\textbf{Non-Geography} & False Positive (\textit{FP}) & True Negative (\textit{TN})  \\
\bottomrule
\end{tabular}
\end{table*}

\section{Experiment}
In this section, we conducted experiments on the Chinese dataset by using the neural network model commonly used in text classification. Through the study of experimental results, we judged that the experimental effect of geoGAT was great.
\subsection{Dataset}

We separately extract data from several commonly used Chinese text classification datasets to form the new dataset.

\textbf{ByteDance\_News}\footnote{https://github.com/fate233/toutiao-text-classfication-dataset}: This dataset is created by ByteDance through the collection of news. The data belongs to the short text, which contains a total of 382,688 samples in 15 categories.

\textbf{THUCNews\_Chinese\_Text}\footnote{http://thuctc.thunlp.org/}: THUCNews filtered and generated historical data from 2005 to 2011 in Sina News RSS subscriptable channel in UTF-8 plain text format\cite{sun2016thuctc}. The whole dataset was divided into 14 categories with a total of 740,000 news documents.

\textbf{Fudan University Chinese Text}: This data is from Fudan University. It is a short text containing a total of 9,804 documents in 20 categories.

\textbf{SogouCA}\cite{wang2008automatic}\footnote{https://www.sogou.com/labs/resource/ca.php}: Sogou Labs collected news data from 18 different domestic and overseas channels from June to July 2012 on a number of news sites.

To build a textual dataset that contains geographic information. In the category of geographical text, we require the geographical noun in the text to have a clear geographical location corresponding.The text can not appear virtual geographical name. The description of a location in each text should be as complete as possible. If words such as directions appear in the text, they need to be accompanied by an exact geographic entity. No English or abbreviations are used to represent geographic information. No Chinese phonetic or abbreviations of geographic names are included. For geographic texts with ambiguities, we also introduce some of these texts, which account for 2\%. The other four categories are games, cars, economy and entertainment.
No other geographic information text is allowed in these four types of text to maintain the integrity of the data. The text with character length less than 200 is first extracted and then the text containing English and Chinese phonetic is filtered by pattern matching method when extracting text. 
A total of 55,000 pieces of data including and excluding geographic information were extracted from the above data to form the corpus. For each category of data, we select evenly from the dataset above. The most important of which was that the geographic category contained 11,000 pieces of data.
The entire dataset has 44,000 as training sets, 11,000 as testing sets, and 4,400 as validation sets. At the end of data extraction, the operation of preprocessing the data.
First, the Chinese stop-words list is used to remove stop-words in the text. Second, special characters, such as special punctuation marks are removed from the text. Although these symbols are very important for human understanding of the text, they cannot be understood by computer classification algorithms.Third, correct spelling and grammar errors in the text.
The entire dataset is divided into two types, one character-level and the other word-level. After data preprocessing, the minimum length, maximum length and average length of the text in the whole dataset are 4,171 and 35.15 respectively.

After data preprocessing is completed, the vocab size of the dataset is 3,776. The actual number of training data is 39,600, which is equal to the number of training set minus the number of validation set. After the text is built into a graph structure, the number of nodes is 58,776, which is equal to the number of text plus vocab size. When calculating the weight of the edge, the number of sliding Windows used is 949,713.
\subsection{Baseline}
We compare several models that performance well in text classification as follows:

\textbf{CNN}: We used CNN-rand, CNN-static and CNN-non-static\cite{kim2014convolutional}. The word vector in CNN-rand was initialized randomly and treated as an optimized parameter in the training process. In CNN-static, the word vector was used Word2Vec\cite{mikolov2013efficient} to get the corresponding result and remained unchanged in the training process. GNN-non-static is similar to the former, except that the word vectors are fine tuned during the training.

\textbf{DCNN}: Dynamic-Convolution-Neural-Networks\cite{kalchbrenner2014convolutional}. The model adopts a dynamic pooling method to conduct semantic modeling for sentences, which is able to solve the problem of variable length of input sentences and is applicable to multiple languages without relying on analytic numbers.

\textbf{LSTM}: RNN is applied in multi-task learning\cite{liu2016recurrent}. Learning by putting multiple tasks together allows information to be shared between multiple tasks.

\textbf{FastText}: This is a word vector and text classification tool. Performance is comparable to deep learning\cite{arm2016bag}. The model input a sub-sequence that the sequence of words and words constitute the feature vector. The feature vector is mapped to the middle layer by a linear transformation and  mapped to the labels. The probability that the final output word sequence belongs to a different category.

\textbf{DPCNN}: A deep CNN at word level\cite{johnson_deep_2017}. The model captured the global semantic representation of the text and gained more performance by increasing the depth of the network without increasing the computational overhead.

\textbf{BiLSTM-Attention}: The attention layer was added to BiLSTM\cite{zhou_attention-based_2016}. BiLSTM used the last time sequence as the output vector, and the attention mechanism weighted all the time sequence vectors as the feature vectors.

\textbf{TextGCN}: It proposes a method that can make the whole corpus into heterogeneous graphs and learn words and document embedding together through GCN\cite{yao_graph_2018}.
\subsection{Model Settings}
In model configuration and data transformation, the embedded dimension of words is set as 300, and each sequence in the training set is represented as a vector, which form is expressed as one-hot encoding, and the whole dataset is represented as a matrix, which is stored as a sparse matrix. Setting the size of the window\_size to 25 to calculate weights between sequence and word and between word and word. 

In the course of training, we set epochs as 1000, the learning rate is set to 0.005, the number of hidden layer units set to 8, the number of head attentions for eight, dropout is set to 0.5, the alpha for leaky\_relu is defined as 0.2, L2 loss on parameters as 5e-4. The slight adjustment of partial parameters has little effect on the experimental results.

\subsection{Evaluation}
The seven most commonly used indicators are used to evaluate the model on the Chinese dataset. The definition of the confusion matrix is shown in Table \ref{confusion matrix}. In this task, the positive example is classified as a geography, and the negative example is classified as a non-geography. $TP$ is the number of positive examples predicted into positive examples. $TN$ is the number of negative examples predicted into negative examples. $FP$ is the number of negative examples predicted to be positive. $FN$ is the number of positive examples predicted into negative examples.

\begin{itemize}
    \item [(1)]
    Recall: Recall is a measure of coverage, measuring how many positive examples are classified as positive examples. This measures the ability of a classifier to recognize positive examples, also known as "sensitivity". 
    
    $\textit{Recall}=\frac{TP}{TP+FN}$.
    \item [(2)]
    Precision: Precision is the number of true positive examples in the sample that predicted positive examples for the prediction results.
    
    $\textit{Precision}=\frac{TP}{TP+FP}$.
    \item [(3)]
    F-score: The relationship between accuracy and recall is opposite. If one goes up, the other goes down. The F-score is a combination of the two.
    
    $\textit{F-score}=2\cdot\frac{Recall\times Precision}{Recall+Precision}$.
    \item [(4)]
    Accuracy: Accuracy is a common and intuitive evaluation index, which is calculated by dividing the number of correctly classified samples by the number of samples.
    
    $\textit{Accuracy}=\frac{TP+TN}{TP+TN+FP+FN}$.
\end{itemize}
We also use macro averaging to evaluate the effectiveness of the model.
Macro averaging are calculated by averaging the precision, recall and F-score for each class.
\begin{itemize}
    \item [(5)]
    Macro-Recall: Macro-Recall is an arithmetic mean of recalls for all classes.
    
    $\textit{Macro-R}=\frac{1}{n}\sum_{i=1}^nR_i$.
    \item [(6)]
    Macro-Precision: Macro-Precision is an arithmetic mean of precisions for all classes.
    
    $\textit{Macro-P}=\frac{1}{n}\sum_{i=1}^nP_i$.
    \item [(7)]
    Macro-F Score: Macro-F Score is an arithmetic mean of F Scores for all classes.
    
    $\textit{Macro-F1}=\frac{2\times P_{macro} \times R_{macro}}{P_{macro} + R_{macro}}$.
\end{itemize}

\subsection{Analysis of Experimental Results}

Table \ref{tab:model effect} describes the indicators of these models in the dataset. In parameter setting, most of the parameters follow the value in the original paper. In order to adapt to the new Chinese dataset in the paper, a small part of parameters have been fine-tuned to get better results. In CNN-rand, the vector of different words is randomly initialized. The pre-trained word vector is used in CNN-static, which is not adjusted during the training process. The pre-trained word vector is also used in CNN-non-static, and then the word vector is fine-tuned. It can be seen from the experimental results that the accuracy of classification is improved by using the pre-trained word vectors, and the experimental results are also slightly improved by fine-tuning word vectors. DCNN introduced one-dimensional convolutional layer and dynamic K-max pooling layers. The size of the middle convolutional layer changes with the length of sentences. It performed better than the traditional convolutional neural networks. Unsupervised embedding layer is introduced as input in DPCNN to further improve the performance of two-view embedding model, which has been significantly improved compared with classic TextCNN. BiLSTM is a variant of RNN. Since RNN has the function of short-term memory, it is suitable for processing sequence problems such as natural language. By introducing gating mechanism, BiLSTM can capture the long distance relation of input samples, which performs well in processing long text. BiLSTM-Attention applied bidirectional LSTM and attention layer. The attention mechanism calculates the weight of each time sequence in LSTM, weights all the time sequences as feature vectors, and then conducts classification. BiLSTM-Attention improves the experimental results of LSTM. FastText is a classification algorithm. This algorithm can speed up the training while maintaining high classification accuracy and train word vectors by itself. geoGAT and TextGCN are graph-based methods to deal with the classification of short texts. Due to the insufficient information of the short texts themselves, additional information is added in the form of graphs to make use of the relationship between samples in corpus for semi-supervised learning. It has a great experimental performance in the short text classification.

Table \ref{tab:macro_avg} describe how each model performs on the macro averaging. CNN-non-static gets better results than CNN-static and CNN-rand by fine-tuning pre-trained word vectors.
From the conclusion Macro-Precision and Macro-F Score, DCNN is slightly inferior to CNN-non-static. In terms of Macro-Recall, DCNN is stronger than traditional TextCNN, and its performance in DPCNN is not as good as TextCNN in this task.
BiLSTM performed about the same as CNN-rand.
BiLSTM-Attention has a better than BiLSTM after introducing the attention mechanism.
For FastText, it is theoretically unnecessary to use the pre-trained word vector. As described in the experimental results, the effect of the pre-trained word vector on the model is improved. Both TextGCN and geoGAT aggregate the characteristics of neighboring vertices onto the central vertex. The reason why the geoGAT works so well is that GCN applied laplace matrix for aggregation while GAT used attention coefficient for aggregation. GAT aggregates the characteristics of neighbor nodes, making the correlation of node characteristics more specific integrated into the model. 

\begin{table*}
  \caption{The performance of different models on the Precision, Recall and F-Score, all of the $p-value<0.05$}
  \label{tab:model effect}
 \scalebox{0.885}{
 \begin{tabular}{lc|cc|cc|cc}
    \toprule
    \multicolumn{2}{c|}{\textbf{Model}}&\multicolumn{2}{c|}{\textbf{Precision}}&\multicolumn{2}{c|}{\textbf{Recall}}&\multicolumn{2}{c}{\textbf{F-Score}}\\
    \midrule
    \textbf{CNN-rand}\cite{kim2014convolutional}          &\textit{(std, p-value)}& 0.8949& \textit{±0.0044, 1.3724e-5} & 0.9207& \textit{±0.0023, 5.4113e-5}& 0.9076& \textit{±0.0054, 5.7121e-4}\\
    \textbf{CNN-static}\cite{kim2014convolutional}        &\textit{(std, p-value)}& 0.9588& \textit{±0.0021, 1.8473e-4} & 0.9537& \textit{±0.0023, 3.7648e-4} & 0.9563& \textit{±0.0038, 4.1625e-4}\\
    \textbf{CNN-non-static}\cite{kim2014convolutional}    &\textit{(std, p-value)}& 0.9607& \textit{±0.0036, 2.7320e-5} & 0.9383& \textit{±0.0084, 4.5238e-4} & 0.9493&\textit{±0.0032, 8.4724e-6}\\
    \textbf{DCNN}\cite{kalchbrenner2014convolutional}     &\textit{(std, p-value)}& 0.9576& \textit{±0.0067, 4.7497e-3} & 0.9362& \textit{±0.0074, 7.3187e-4} & 0.9467& \textit{±0.0051, 6.7754e-3}\\
    \textbf{BiLSTM}\cite{liu2016recurrent}                &\textit{(std, p-value)}& 0.9418& \textit{±0.0107, 7.2728e-4} & 0.9524& \textit{±0.0082, 9.1703e-5} & 0.9471& \textit{±0.0040, 3.6479e-6}\\
    \textbf{FastText}\cite{arm2016bag}                    &\textit{(std, p-value)}& 0.9648& \textit{±0.0053, 2.9383e-4} & 0.9603& \textit{±0.0029, 3.0656e-5} & 0.9626& \textit{±0.0025, 2.8302e-5}\\
    \textbf{FastText(pre-trained)}&\textit{(std, p-value)}& 0.9688& \textit{±0.0097, 4.2189e-3} & 0.9594& \textit{±0.0032, 3.2164e-5} & 0.9640& \textit{±0.0036, 7.8231e-5}\\
    \textbf{DPCNN}\cite{johnson_deep_2017}                &\textit{(std, p-value)}& 0.9618& \textit{±0.0131, 5.1234e-5} & 0.9478& \textit{±0.0070, 5.1234e-5} & 0.9547& \textit{±0.0038, 5.1234e-5}\\
    \textbf{BiLSTM-Att}\cite{zhou_attention-based_2016}   &\textit{(std, p-value)}& 0.9540& \textit{±0.0075, 1.5507e-2} & 0.9465& \textit{±0.0046, 8.9944e-5} & 0.9502& \textit{±0.0022, 1.8904e-5}\\
    \textbf{TextGCN}\cite{yao_graph_2018}                 &\textit{(std, p-value)}& 0.9708& \textit{±0.0086, 7.6531e-3}& 0.9756& \textit{±0.0058, 1.9470e-3} & 0.9732& \textit{±0.0072, 4.2623e-3}\\
    \textbf{geoGAT}                                       &\textit{(std)}&\textbf{0.9834}&\textit{±0.0044}&\textbf{0.9787}&\textit{±0.0039}&\textbf{0.9810}&\textit{±0.0024}\\
  \bottomrule
\end{tabular}}
\end{table*}
\begin{table*}
  \caption{The performance of each model on the Macro average based on the t-test $(p-value<0.05)$.}
  \label{tab:macro_avg}
  \scalebox{0.885}{
  \begin{tabular}{lc|cc|cc|cc}
    \toprule
    \multicolumn{2}{c|}{\textbf{Model}}&\multicolumn{2}{c|}{\textbf{Macro-Precision}}&\multicolumn{2}{c|}{\textbf{Macro-Recall}}&\multicolumn{2}{c}{\textbf{Macro-F Score}}\\
    \midrule
    \textbf{CNN-rand}\cite{kim2014convolutional}        &\textit{(std, p-value)}& 0.9022 &\textit{±0.0012, 9.1844e-6}& 0.9043&\textit{±0.0023, 8.7348e-5}& 0.9020&\textit{±0.0067, 4.3742e-6} \\
    \textbf{CNN-static}\cite{kim2014convolutional}      &\textit{(std, p-value)}& 0.9203 &\textit{±0.0097, 1.7253e-3}& 0.9185&\textit{±0.0077, 1.8431e-2} & 0.9223&\textit{±0.0089, 1.8734e-3} \\
    \textbf{CNN-non-static}\cite{kim2014convolutional}  &\textit{(std, p-value)}& 0.9207 &\textit{±0.0014, 7.5254e-6}& 0.9205&\textit{±0.0013, 5.1784e-6} & 0.9205&\textit{±0.0014, 2.4123e-6} \\
    \textbf{DCNN}\cite{kalchbrenner2014convolutional}   &\textit{(std, p-value)}& 0.9130 &\textit{±0.0027, 5.2944e-4}& 0.9232&\textit{±0.0019, 1.2987e-6} & 0.9139&\textit{±0.0016, 3.0083e-3}\\
    \textbf{BiLSTM}\cite{liu2016recurrent}              &\textit{(std, p-value)}& 0.9096 &\textit{±0.0035, 1.0788e-6}& 0.9086&\textit{±0.0036, 9.3625e-7} & 0.9090&\textit{±0.0040, 4.9350e-7} \\
    \textbf{FastText}\cite{arm2016bag}                  &\textit{(std, p-value)}& 0.9257 &\textit{±0.0028, 3.1640e-5}& 0.9258&\textit{±0.0021, 2.6435e-5} & 0.9257&\textit{±0.0020, 1.4186e-5} \\
    \textbf{FastText(pre-trained)}                      &\textit{(std, p-value)}& 0.9266 &\textit{±0.0017, 3.0737e-5}& 0.9354&\textit{±0.0045, 1.2872e-4} & 0.9287&\textit{±0.0021, 1.3885e-5} \\
    \textbf{DPCNN} \cite{johnson_deep_2017}             &\textit{(std, p-value)}& 0.9164 &\textit{±0.0013, 2.0790e-6}& 0.9156&\textit{±0.0020, 1.7827e-6} & 0.9155&\textit{±0.0015, 6.1659e-7} \\
    \textbf{BiLSTM-Att}\cite{zhou_attention-based_2016} &\textit{(std, p-value)}& 0.9076 &\textit{±0.0028, 1.5900e-6}& 0.9066&\textit{±0.0033, 1.2662e-6}& 0.9064&\textit{±0.0032, 6.5985e-7} \\
    \textbf{TextGCN}\cite{yao_graph_2018}               &\textit{(std, p-value)}& \textbf{0.9512} &\textit{±0.0016, 1.9558e-4}& 0.9507&\textit{±0.0012, 2.9469e-2} & 0.9509&\textit{±0.0019, 4.8870e-3} \\
    \textbf{geoGAT}                                     &\textit{(std)}&0.9510&\textit{±0.0059}&\textbf{0.9524}&\textit{±0.0047}&\textbf{0.9520}&\textit{±0.0049}\\
  \bottomrule
\end{tabular}}
\end{table*}
\subsection{Analysis of Word Segmentation}
Chinese words are not separated by spaces like English and there is no separator between words and words. The meaning group of the text is difficult to distinguish so the way of word segmentation will have an impact on the training of the model.

\begin{figure}
    \centering
    \includegraphics[width=1\textwidth]{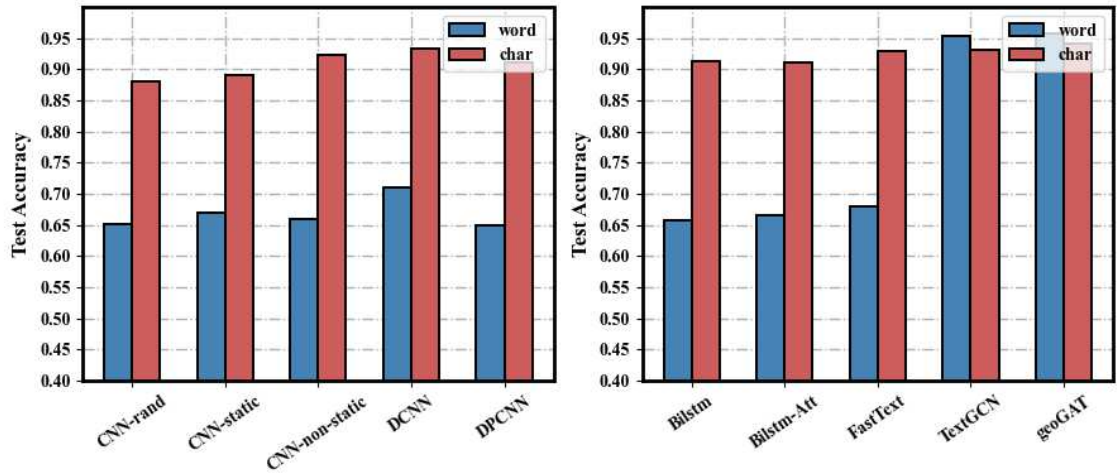}
    \caption{Test accuracy of the models}
    \label{fig:word-char1}
\end{figure}

We used two ways of word segmentation to train the model.
The first is char-level, which separates each word in each piece of data by a space.
The second is word-level, which divides each piece of data using the Jieba word segmentation tool. Figure \ref{fig:word-char1} show the result of different models using different segmentation methods on test accuracy. In the bar chart, blue represents word level segmentation and red represents character level segmentation. It can be concluded from the bar graph that, in addition to the graph neural network, the effect of other models based on CNN and RNN is greatly affected by the word segmentation mode and the model trained with char-level word segmentation has a good performance. 

There are several reasons for the poor performance of the word-level model.
The first reason is that in the dataset we used, the amount of data is not large. Each piece of data is not long. It belongs to the short text. Therefore, after the data is segmented, the frequency of each word is very small, which leads to the problem of data sparsity. Overfitting is easy to occur during model training, thus limiting the learning ability of the model. The second reason is that word segmentation sometimes introduce noise into the original data. Due to the limitations of word segmentation tools and algorithms, there is no way to achieve the accuracy as humans do.
The third reason is that the benefits brought by word segmentation are not obvious. Word segmentation reduce the polysemy of the word to a certain extent so that the information carried by the word is more abundant than the word.
But neural networks can learn words and combinations of words through a complex learning process.

As Figure \ref{fig:word-char1} shows, whether or not to use word-level segmentation TextGCN and geoGAT has little effect on how well the model is classified.
The reason why is that during the training, the entire dataset is first constructed into a heterogeneous graph, the sequence relationship between sentences is connected through the graph structure and the weights of nodes and edges of the graph strengthen the connection between the texts. Whether each node contains a word or a word does not affect the amount of information contained in the entire graph structure. One-hot encoding is used as the feature input, and although the dimension of the encoding is high, this approach almost never loses information between texts. 
\begin{figure}
    \centering
    \includegraphics[width=1\textwidth]{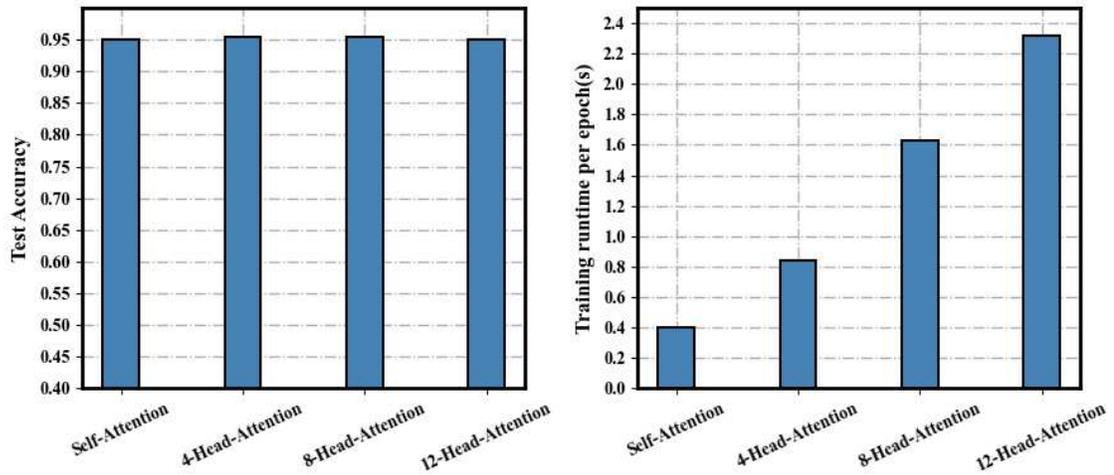}
    \caption{The impact of the amount of head attention on test accuracy and runtime}
    \label{fig:num-head-test-accuracy}
\end{figure}

\subsection{Ablation Studies}
In order to stabilize the learning process of self-attention, geoGAT adopts Multi-head attention.
Multi-head attention is actually a combination of multiple self-attention structures.
Each head learn the features in different representation spaces, so the focus of attention learned by multiple heads is different, which enables the model to have a larger capacity. The amount of head-attention also influence the performance of the model to some extent.

\textbf{Training runtime and test accuracy of multi-head attention and self-attention}. As shown in the left side of Figure \ref{fig:num-head-test-accuracy}, the x-axis is the amount of head attention, and the y-axis is the test accuracy.
In the right side of Figure \ref{fig:num-head-test-accuracy}, the x-axis is the same as the former, and the y-axis is the training runtime per epoch. We tested the training running time and final test accuracy of the model in each epoch under the conditions of self-attention, 4-head attention, 8-head attention and 12-head attention respectively. The experimental results show that as the amount of head-attention increases, so does the time spent on model training.
With more head-attention, the calculation amount of output features of each node of the model will increase, and the training time will increase.

In terms of test accuracy, self-attention performed worse than multi-head attention.
For multi-head attention, the amount of head-attention has no impact on the test results. A proper amount of head attention can learn enough features but too much of it is a waste of computing resources
Moreover, the experimental results also indicate that more head-attention slows down the training speed of the model.

\begin{figure}
\subfigure[]
  {
    \includegraphics[width=0.485\textwidth]{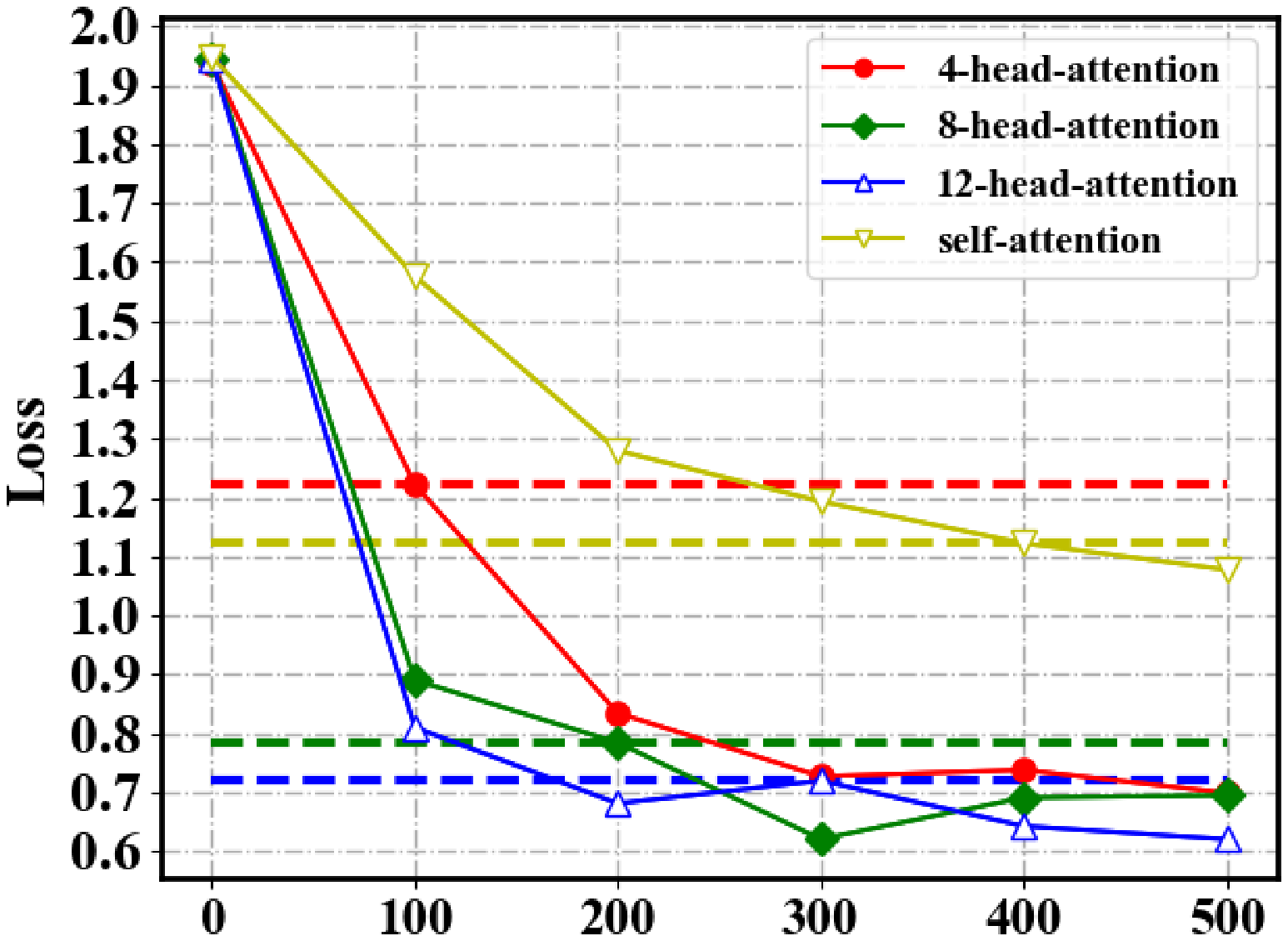}
    \label{fig:subfig:loss-epoch}
  } 
\subfigure[]
  {  
    \includegraphics[width=0.485\textwidth]{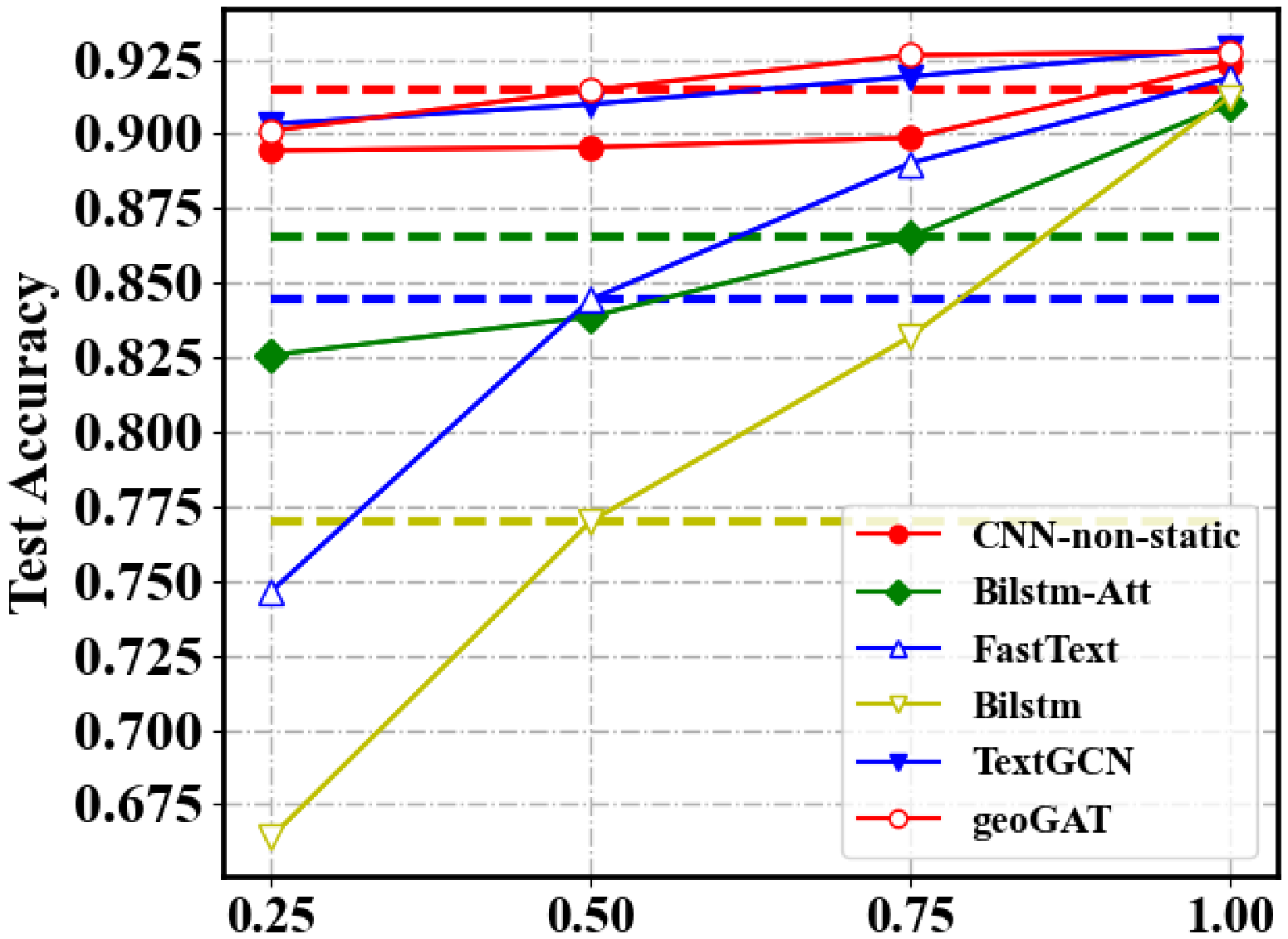}
    \label{fig:subfig:train-rate}
  }
\caption{The performance of loss in different number of head attention and Test accuracy on various labeled data proportions
}
\label{fig:loss-epoch-train-rate}
\end{figure}
\textbf{Loss of multi-head attention and self-attention}. It also has an impact on the convergence speed of loss. As shown in Figure \ref{fig:subfig:loss-epoch}, the x-axis in the figure represents the number of epochs, while the y-axis is the value of loss. As far as self-attention is concerned, the decrease amplitude of loss was the smallest after the model trained an epoch of the same size.
With the increase in the amount of head-attention, loss decreases gradually.
After the training of 200 epochs, the loss in multi-head attention reached convergence almost at the same time, and the subsequent changes were small.

\textbf{Different proportion of labeled data on test accuracy}. On account of the large data traffic, we have acquired a great many of texts containing geographical knowledge, so the use of deep learning to separate these texts requires a large amount of data annotation work.
If the model can achieve high accuracy in classified data through a small amount of labeled data, it is necessary to develop weakly supervised learning.
As shown in Figure \ref{fig:subfig:train-rate}, the experiment trained several models by reducing the amount of training data with labels, and obtained their changes in test accuracy.
The x-axis in the figure represents the proportion with labels in the data, while the ordinate represents the accuracy rate. 

As can be obtain from the figure, BiLSTM was the most significantly affected by data changes. By increasing the proportion of labeled data, the accuracy rate was increased from 60$\%$ to 90$\%$, followed by FastText, which was increased from 70$\%$ to 90$\%$.
After the attention mechanism is introduced, the range of change of BiLSTM decreases.
Therefore, the attention mechanism enables BiLSTM to learn the features of text more emphatically with a small amount of annotation data, so as to achieve better classification.
TextGCN and geoGAT have a negligible effect by changes in labeled data. GCN is a calculation method of the whole graph and the node characteristics of the whole graph will be updated once the calculation is done. The learned parameters are closely related to the graph mechanism.

For geoGAT the edges to reinforce the importance of model by learning performance, there are essentially two operation ways. Global graph is attention to each vertex in the graph of calculation at arbitrary node on the attention, the structure of this method does not rely on chart, suitable for inductive task. However, the second method is the commonly used mask graph attention, which only calculates the first-order neighbor node's attention and used multi-head attention. Although the number of labels is reduced, the features between nodes can still be learned from other nodes.
\section{DISCUSSION AND FUTURE WORK}
Our research problem is to separate the text containing geographic information from the large number of network text.
In our experiment, our method has a good performance on the classification of Chinese geographical text. This method is based on the transformation of text into heterogeneous graph, and the transformation of Chinese text into graph structure through text processing. The model method is a graph neural network built on the basis of attention, which obtain features by calculating the attention of adjacent nodes. A small amount of label data can also achieve good results in the model.
The experimental results show that geographic text classification is of great necessity and importance, and it is significant to further extract knowledge from geographic text.

There are some limitations in this study.
In the process of classification, some texts contain ambiguous data, while others may contain a small amount of geographic information, resulting in misidentification. The extracted geographic information text is not enough for geolocation. Increasing the number of layers in the graph attention network is difficult and the training method is complicated.

In the future work, we hope to improve the accuracy of classification by means of entity disambiguation and to extract as detailed a geographic location as possible from text with geographic information.
With reference to model building, the training way of geoGAT can also be optimized.
At present, geoGAT is applied in the network of single-layer graph structure. If it is extended to multi-layer network can have better computing effect.

\section{CONCLUSION}
We introduce a method to classify geographic texts.
The geoGAT model uses the form of transforming text into heterogeneous graphs, and introduces the attention mechanism to effectively combine the features in the graph to apply to geographic Chinese text datasets.

After experiments on multiple baseline models, the model we proposed has great performance in the classification of geographic texts by combining the features of graphs and nodes. The advantages of graph neural network in network data processing are verified again.

\begin{acks}
The work described in this paper is supported by National Natural Science Foundation of China (31770768), Fundamental Research Funds for the Central Universities(2572017PZ04), Heilongjiang Province Applied Technology Research and Development Program Major Project(GA18B301,GA20A301) and China State Forestry Administration Forestry Industry Public Welfare Project (201504307). We would like to thank all the reviewers for all insightful comments and suggestions.
\end{acks}


\bibliographystyle{ACM-Reference-Format}
\bibliography{geoGAT}

\end{document}